\documentclass[letterpaper, 10 pt, conference]{ieeeconf}  

\IEEEoverridecommandlockouts                              

\usepackage{amsmath,amsfonts}
\usepackage{algorithmic}
\usepackage{subfigure}
\usepackage{subcaption}
\usepackage{multirow, booktabs}
\usepackage{graphicx}
\usepackage{adjustbox}

\title{\LARGE \bf
TrajMamba: An Ego-Motion-Guided Mamba Model for Pedestrian Trajectory Prediction from an Egocentric Perspective
}

\author{Yusheng Peng$^{1}$, Gaofeng Zhang$^{2}$ and Liping Zheng$^{2*}$
\thanks{*This work was supported in part by the National Natural Science Foundation of China under Grant 62372152, in part by the Fundamental Research Funds for the Central Universities of China under Grant JZ2023HGQB0481, and in part by China Postdoctoral Science Foundation under Grant 2023M740961}
\thanks{$^{1}$Yusheng Peng is with the School of Computer Science and Information Engineering, Hefei University of Technology, Hefei 230601, China (e-mail: wisionpeng@hfut.edu.cn).}
\thanks{$^{2}$Gaofeng Zhang and Liping Zheng are with the School of Software, Hefei University of Technology, Hefei 230601, China (e-mail: g.zhang@hfut.edu.cn, zhenglp@hfut.edu.cn).}
\thanks{$^{*}$Corresponding author}
}

\begin{document}

\maketitle
\thispagestyle{empty}
\pagestyle{empty}

\begin{abstract}

Future trajectory prediction of a tracked pedestrian from an egocentric perspective is a key task in areas such as autonomous driving and robot navigation. The challenge of this task lies in the complex dynamic relative motion between the ego-camera and the tracked pedestrian. To address this challenge, we propose an ego-motion-guided trajectory prediction network based on the Mamba model. Firstly, two Mamba models are used as encoders to extract pedestrian motion and ego-motion features from pedestrian movement and ego-vehicle movement, respectively. Then, an ego-motion-guided Mamba decoder that explicitly models the relative motion between the pedestrian and the vehicle by integrating pedestrian motion features as historical context with ego-motion features as guiding cues to capture decoded features. Finally, the future trajectory is generated from the decoded features corresponding to the future timestamps. Extensive experiments demonstrate the effectiveness of the proposed model, which achieves state-of-the-art performance on the PIE and JAAD datasets.

\end{abstract}

\section{Introduction}
Predicting future trajectories of pedestrians is a key aspect of pedestrian behavior analysis, which is essential for ensuring the safe driving of autonomous vehicles and the secure navigation of mobile robots in dynamic crowd environments. Most research \cite{Wang2024Pedestrian,Lin2024DyHGDAT} model and analyze pedestrian movement from a bird's-eye view to predict the future trajectories of pedestrians. This type of method is only suitable for third-person perspective videos captured by surveillance cameras or drones. Autonomous vehicles and mobile robots typically perceive their surroundings from a first-person perspective. Specifically, these ego cameras usually move in tandem with the terminal agent. Nevertheless, the dual motion of the ego camera and the tracked pedestrian poses significant challenges in predicting the pedestrian's future trajectory from the egocentric perspective.

The core objective of first-person perspective pedestrian trajectory prediction is to accurately forecast the future bounding box sequence of the target pedestrian within the ego-camera coordinate system. The primary challenge stems from the fact that the apparent motion observed in the image is the result of the superimposed projection of both the pedestrian's intrinsic motion and the carrier's (vehicle or camera) motion onto the two-dimensional imaging plane. Consequently, the prediction accuracy heavily relies on the effective modeling of the relative motion between these two sources. Most existing approaches tackle this challenge through feature-level fusion strategies, where pedestrian motion and vehicle ego-motion are combined in the latent space using early or late fusion techniques, and subsequently fed into a decoder for end-to-end prediction. For example, methods \cite{Rasouli2021Bifold,Feng2021Using} employ LSTM networks as encoders to extract pedestrian motion features and vehicle ego-motion features separately, and then concatenate these two feature types as input to the decoder LSTM. Similarly, method \cite{Zhang2024Decouple} adopts Transformer to replace LSTM for both encoding and decoding. Meanwhile, method \cite{Damirchi2023} concatenates the two types of features at the embedding level and processes them through a Transformer to obtain fused encoded representations, which are then decoded to generate future trajectories. In all these approaches, the two types of motion features are integrated into a unified representation during the encoding stage, leaving the decoder to generate future predictions. This technical paradigm results in a rather ambiguous modeling of the relative motion relationship between the two motion sources, making it difficult to explicitly analyze how ego-motion dynamically regulates pedestrian movement. 

To address these limitations, we propose TrajMamba, a novel ego-motion-guided framework for egocentric pedestrian trajectory prediction. Specifically, we first employ two Mamba encoders to extract pedestrian motion features and ego-motion features, respectively. An ego-motion-guided Mamba decoder is then designed, which takes pedestrian motion features as historical context and ego-motion features as future guiding cues. By leveraging Mamba's sequential modeling capacity, the decoder explicitly captures the dynamic modulation mechanism of ego-motion on pedestrian movement to infer future motion features. Finally, these decoded features are mapped to future trajectories via a prediction head. The main contributions of this work are as follows:

\begin{itemize}
	\item We propose TrajMamba, a novel ego-motion-guided framework based on the Mamba architecture, for predicting future pedestrian trajectories from an egocentric perspective. 
	\item We design an ego-motion-guided Mamba decoder that explicitly models the relative motion between the pedestrian and the camera by integrating pedestrian motion features as historical context with ego-motion features as guiding cues for future prediction.
	\item Extensive experiments on the PIE and JAAD datasets demonstrate that TrajMamba outperforms baseline methods on the ADE, FDE, ARB, and FRB metrics.
\end{itemize}

\section{Related Work}
\subsection{Pedestrian Trajectory Prediction}
Current methods for predicting pedestrian trajectories are classified into those based on the bird's eye perspective and those based on the egocentric perspective. Bird's-eye perspective-based methods focuses on predicting the 2D coordinates of pedestrians' future locations in bird's eye videos captured by surveillance cameras or drones. As a typical sequence prediction task, many works \cite{Chen2024Goal,ZhouEdge2025} employ the recurrent neural networks (RNNs) or its variants (LSTM and GRU) to capture temporal dependencies of pedestrian's movement, and utilize pooling mechanism \cite{Liang2021Temporal} or graph neural networks \cite{Wang2023SEEM} to model spatial interactions among pedestrians. Recent works \cite{Zhou2025Heterogeneous,Chen2024Stochastic} have utilized Transformers to replace networks such as LSTM, leveraging their attention mechanisms to model temporal dependencies of pedestrian movement and spatial interactions. Additionally, some researchers \cite{Mao2023Leapfrog,Wu2024Motion} integrate recently popular diffusion models and introduce new architectures to predict pedestrian trajectories. Moreover, some works \cite{Guo2022End,Zhang2025Group} leverage the visual information to improve the prediction performance.  

On the other hand, egocentric perspective-based methods focus on predicting future bounding boxes of tracked pedestrians from egocentric videos captured by vehicle-mounted or wearable cameras. Inspired by bird's-eye view future localization methods, some researchers \cite{Bhattacharyya2018Long,Quan2021Holistic,Huynh2023Online} use LSTM or GRU to process the bounding boxes of tracked pedestrians to model their motion dynamics and predict the bounding boxes of future locations. Furthermore, the researchers \cite{Malla2020TITAN,Czech2022On} thoroughly investigate pedestrian behavior attributes, incorporating pose or behavior labels as auxiliary information in their models, which enhances the accuracy of future localization. Moreover, some methods \cite{Rasouli2023PedFormer,Zhang2024Decouple} also incorporate scene semantics to more accurately predict the future locations of pedestrians in diverse environments. The egocentric view is unique in that the camera itself is in motion, presenting new challenges for predicting a pedestrian's future location. To tackle this, researchers \cite{Qiu2021Indoor,Feng2021Using} extract the camera's ego-motion features using the vehicle's travel data or the camera's IMU signals. By combining the pedestrian's motion with the camera's ego-motion, they enhance the accuracy of future localization predictions. Different from the above works, we innovatively introduce the Mamba model for pedestrian motion modeling and ego-motion modeling, which improves the efficiency while maintaining the modeling ability.

\subsection{State Space Models}
Recently, state-space models (SSMs) \cite{Gu2022Efficiently,Gupta2022Diagonal} have garnered significant attention due to their excellent performance in sequence modeling and inference. Mamba \cite{Gu2023Mamba} introduces a selective state space model architecture that integrates time-varying parameters into the SSM framework and proposes a hardware-aware algorithm to enhance the efficiency of training and inference processes. Furthermore, Mamba-2 \cite{Dao2024Transformers} refines this by linking SSMs to attention variants, achieving 2-8x speedup and performance comparable to transformers. SSMs and Mamba models are widely and successfully used in many tasks. For instance, Zhu et al. \cite{Zhu2024Vision}  introduce a new generic vision backbone called Vision Mamba (Vim), which incorporates bidirectional Mamba blocks. Vim leverages position embeddings to mark image sequences and compresses visual representations using bidirectional state-space models. Similarly, Park et al. \cite{Park2024VideoMamba} introduce VideoMamba for video analysis, which tackles the unique challenge of integrating non-sequential spatial with sequential temporal information in video processing through spatio-temporal forward and backward SSM. Tang et al.\cite{Tang2025Spatial} propose a novel spatial-temporal graph Mamba (STG-Mamba) for the music-guided dance video synthesis task, i.e., to translate the input music to a dance video. He et al. \cite{He2025Decomposed} propose a decomposed spatio-temporal Mamba (DST-Mamba) for traffic prediction. Zhang et al. \cite{Zhang2025Point} introduce a Mamba-based point cloud network named Point Cloud Mamba, which incorporates several novel techniques to help Mamba better model point cloud data. Inspired by these, we explore a novel application of the Mamba model for pedestrian trajectory prediction from an egocentric perspective.

\begin{figure*}[!t]
	\centering
	\includegraphics[scale=0.53]{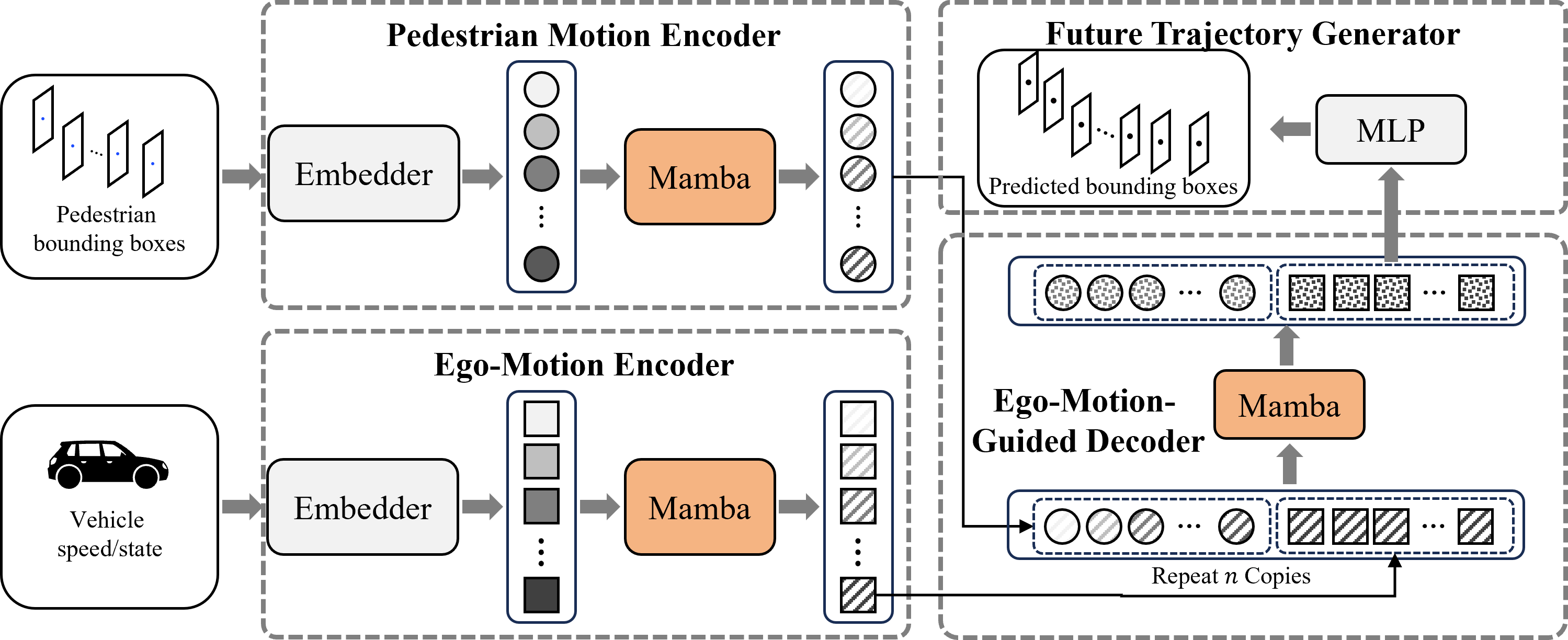}
	\caption{Overview of the proposed TrajMamba.}
	\label{fig:overview}
\end{figure*} 

\section{Method}

\subsection{Overview}
Our goal is to forecast the future bounding boxes of pedestrians from an egocentric perspective. In this section, we propose a novel ego-motion-guided Mamba model named TrajMamba. As shown in Fig. \ref{fig:overview}, the proposed TrajMamba model includes four modules: pedestrian motion encoder (PME), ego-motion encoder (EME), ego-motion guide decoder (EMGD), and future trajectory generator (FTG) module. Firstly, the pedestrian motion features and the ego-motion features are extracted through the PME and EME modules, respectively. Then, the pedestrian's motion features and ego-motion features are sent to the EMGD module to output the decoded features by modeling the relative motion relationship between the pedestrian and the ego-vehicle. Finally, the decoded features of future timestamps are then used to generate future trajectory through the FTG module. 

\subsection{Problem Formulation}
Suppose that the past bounding boxes of the tracked pedestrian at current frame $t$ are denoted as a sequence $B_{past} = (B_{T-m},\cdots, B_T)$, and its future bounding boxes are denoted as a sequence $B_{future} = (B_{T},\cdots, B_{T+n})$, where $B_t=(x_t, y_t, w_t, h_t)$, and $m$ and $n$ are the lengths of past and future timestamps. The movement information of the ego-vehicle is represented as $V_{past}=(v_{T-m}, \cdots, v_T)$. Our goal is to predict the future bounding boxes $\tilde{B}_{future} = (\tilde{B}_{T},\cdots, \tilde{B}_{T+n})$ of the tracked pedestrian from its past bounding boxes $B_{past}$ and the ego-vehicle's past movement information $V_{past}$.

\subsection{Input and Output Representation}
Existing approaches to pedestrian motion representation can be categorized into two paradigms: one utilizes the center point coordinates with bounding box width and height \cite{Huynh2023Online, Yang2025Interact}, while the other employs the top-left and bottom-right corner coordinates \cite{Zhang2024Decouple, Hu2025Probabilistic}. Both paradigms typically maintain consistent input-output representations. Recently, differentiated representation strategies have emerged. For example, method \cite{Rasouli2023PedFormer} inputs corner coordinates and velocities but outputs regressed corner-based bounding boxes. Method \cite{Liu2024Egocentric} inputs center coordinates with velocity and acceleration, yet outputs only the predicted center coordinates. Motivated by this, we propose a differentiated motion representation method. Specifically, given a pedestrian's historical bounding box sequence, we compute the center point displacement (velocity) and scale variation between adjacent frames as follows:
\begin{gather}
	v_t^x=x_t-x_{t-1}, v_t^y=y_t-y_{t-1} \\
	\Delta w = w_t-w_{t-1}, \Delta h=h_t-h_{t-1}
\end{gather} 
Based on this, the motion state of the pedestrian at time step t is constructed as $M_t=(x_t,y_t,w_t,h_t,v_t^x,v_t^y,\Delta w,\Delta h)$.

On the other hand, inspired by the constant velocity assumption in \cite{Style2019Forecasting}, we introduce the Constant Velocity and Constant Scaling (CV-CS) assumption. The key innovation lies in predicting residuals relative to a physically motivated reference trajectory, which not only eases network optimization but also injects valuable prior knowledge into the learning process. Specifically, based on the CV-CS assumption, pedestrian motion approximately maintains constant velocity and constant scale over short time intervals. Therefore, we compute the average velocity $(\bar{v}^x, \bar{v}^y)$ and average scale variation rate $ (\bar{h},\bar{h})$ using the last five frames of historical observations. According to this assumption, the reference bounding box at any future time step $t+\tau$ can be inferred:
\begin{gather}
	\hat{x}_{t+\tau}=x_T+\bar{v}^x\cdot \tau, \hat{y}_{t+\tau}=y_T+\bar{v}^y\cdot \tau \\
	\hat{w}_{t+\tau}=w_T\cdot(1+\bar{w})^{\tau}, \hat{h}_{t+\tau}=h_t\cdot(1+\bar{h})^{\tau}
\end{gather}
We define the residual offset $\Delta _{T+\tau}$ of the ground-truth future bounding box $ B_{T+\tau}$ relative to the reference $\hat{B}_{T+\tau}$. The network is tasked with predicting this offset, denoted as $\hat{\Delta}_{T+\tau}$. Finally, the predicted bounding box $\tilde{B}_{T+\tau}$ is obtained by adding the predicted offset $\hat{\Delta}_{T+\tau}$ to the reference $\hat{B}_{T+\tau}$. 

This method reformulates the prediction target into a residual form, which not only preserves the physical prior of the CV-CS assumption but also allows the network to flexibly correct deviations from actual motion. Consequently, it achieves more accurate trajectory prediction in complex dynamic scenarios.

\subsection{Pedestrian Motion Encoder} 
To precisely predict the future trajectory of the tracked pedestrian, it is of vital importance to initially model and analyze their past movement dynamics and patterns so as to capture the underlying principles of their behavior. For a pedestrian's past bounding boxes $B_{past}$, we begin by calculating the motion state $M_t=(x_t,y_t,w_t,h_t,v_t^x,v_t^y,\Delta w,\Delta h)$. Then, we utilize a multilayer perceptron (MLP) as the embedder layer to transform the motion state sequence $\{M_{T-m},\cdots,M_T\}$ into embedding features $ E_{pm} \in \mathbb{R} ^{(T-m)\times D}$. The $D$ is the dimension of each embedding feature. Existing methods always employ LSTM \cite{Yao2021BiTrap}, GRU \cite{Wang2022Stepwise}, or Transformer \cite{Zhang2024Decouple} model as the encoder to capture motion features. Different from the previous methods, we use the Mamba model as the encoder, aiming to effectively extract motion features and improve efficiency by taking advantage of its linear time series modeling ability and efficient long sequence processing performance. The encode operation of the Mamba model is formulated as follows:
\begin{equation}
	F_{pm}=Mamba(E_{pm})
	\label{eq:5}
\end{equation}

\subsection{Ego-Motion Encoder} 
Modeling the motion of the ego camera is key to understanding the motion dynamics of the tracked pedestrian in an egocentric perspective. Existing works \cite{Yagi2018Future,Yao2019Egocentric,Neumann2021Pedestrian} utilize CNNs to implicitly capture ego-motion features from image streams. Other works usually use LSTMs \cite{Bhattacharyya2018Long,Rasouli2021Bifold} or Transformers \cite{Yin2021Multimodal,Zhang2024Decouple} to model the movement information of ego vehicles to capture ego motion features. Inspired by this, we use a Mamba model to extract the motion features of ego vehicles. Specifically, given the speed or state information $v_t$ of the vehicle at $t$ timestamp, an MLP layer is used as an embedder to transform this information into an embedded feature $e_t$ and obtain the embedded feature sequence $E_{em}=\{e_{t-m},...,e_{t}\}$ of past $m$ timestamps. Then, a Mamba model is used as an encoder to transform the embedded features $E_{em}$ into ego-motion features $F_{em}$ through sequential modeling. The encode operation of the Mamba model is formulated as follows:
\begin{equation}
	F_{em}=Mamba(E_{em})
	\label{eq:6}
\end{equation}

\begin{table*}[!t]
	\centering
	\caption{Performance of the proposed TrajMamba and other existing models on the PIE and JAAD datasets.}
	\begin{tabular}{ccccccccccc}
		\toprule
		\multirow{3}[6]{*}{Models} & \multirow{3}[6]{*}{Publications} & \multirow{3}[6]{*}{Years} & \multicolumn{4}{c}{PIE}       & \multicolumn{4}{c}{JAAD} \\
		\cmidrule{4-11}          &       &       & ADE $\downarrow$  & FDE $\downarrow$  & ARB $\downarrow$  & FRB $\downarrow$  & ADE $\downarrow$  & FDE $\downarrow$  & ARB $\downarrow$  & FRB $\downarrow$ \\
		\cmidrule{4-11}          &       &       & \multicolumn{8}{c}{Observation frames: 15 (0.5s)  Prediction frames: 30 (1.0s)} \\
		\midrule
		FOL \cite{Yao2019Egocentric}   & ICRA  & 2019  & 73.87 & 164.53 & 78.16 & 143.49 & 61.39 & 126.97 & 70.12 & 129.17 \\
		FPL \cite{Yagi2018Future}   & CVPR  & 2018  & 56.66 & 132.23 & /     & /     & 42.24 & 86.13 & /     & / \\
		B-LSTM \cite{Bhattacharyya2018Long} & CVPR  & 2018  & 27.09 & 66.74 & 37.41 & 75.87 & 28.36 & 70.22 & 39.14 & 79.66 \\
		PIE\_traj \cite{Rasouli2019PIE} & ICCV  & 2019  & 21.82 & 53.63 & 27.16 & 55.39 & 23.49 & 50.18 & 30.40  & 57.17 \\
		PIE\_full \cite{Rasouli2019PIE} & ICCV  & 2019  & 19.50  & 45.27 & 24.40  & 49.09 & 22.83 & 49.44 & 29.52 & 55.43 \\
		BiPed \cite{Rasouli2021Bifold} & ICCV  & 2021  & 15.21 & 35.03 & 19.62 & 39.12 & 20.58 & 46.85 & 27.98 & 55.07 \\
		PedFormer \cite{Rasouli2023PedFormer} & ICRA  & 2023  & 13.08 & 30.35 & 15.27 & 32.79 & 17.89 & 41.63 & 24.56 & 48.82 \\
		BiTrap \cite{Yao2021BiTrap} & RAL   & 2021  & 8.83  & 18.52 & 12.61 & 22.97 & 19.02 & 39.28 & 22.57 & 40.88 \\
		SGNet \cite{Wang2022Stepwise} & RAL   & 2022  & 8.35  & 17.83 & 12.32 & 22.58 & 14.90  & 30.88 & 19.19 & 34.80 \\
		\textbf{Ours} & \textbf{/} & \textbf{/} & \textbf{7.72} & \textbf{16.99} & \textbf{11.60} & \textbf{22.00} & \textbf{13.90} & \textbf{30.76} & \textbf{18.04} & \textbf{34.79} \\
		\midrule
		\multicolumn{3}{c}{}  & \multicolumn{8}{c}{Observation frames: 15 (0.5s)  Prediction frames: 45 (1.5s)} \\
		\midrule
		PEvT \cite{Neumann2021Pedestrian} & CVPR  & 2021  & 19.15 & 45.98 & /     & /     & 21.08 & 49.08 & /     & / \\
		MTN \cite{Yin2021Multimodal}  & IJCAI & 2021  & 18.89 & 45.50  & /     & /     & 20.90  & 48.55 & /     & / \\
		Zhang et al. \cite{Zhang2024Decouple} & TIP   & 2024  & 17.41 & 44.92 & /     & /     & \textbf{17.92} & \textbf{41.33} & /     & / \\
		\midrule
		\textbf{Ours} & \textbf{/} & \textbf{/} & \textbf{13.01} & \textbf{31.08} & \textbf{20.63} & \textbf{41.79} & 23.39 & 55.11 & 32.06 & 63.65 \\
		\bottomrule
	\end{tabular}%
	\label{tab:1}%
\end{table*}%

\subsection{Ego-Motion-Guided Decoder} 

Existing methods primarily integrate pedestrian motion and vehicle ego-motion into a unified encoded feature through early \cite{Damirchi2023}, late \cite{Rasouli2021Bifold,Quan2021Holistic}, or cross \cite{Su2022Crossmodal,Rasouli2024A} fusion strategies, and then generate future trajectories by decoding this fused representation. Although such schemes achieve collaborative modeling of the two motion sources, they struggle to explicitly analyze the dynamic modulation mechanism of ego-motion on pedestrian movement. To overcome the limitation of implicit fusion, we introduce an explicit modeling framework. In this framework, pedestrian motion features are treated as historical context, and ego-motion features are leveraged as explicit conditional cues for future prediction. By employing sequential modeling within the decoder, the network explicitly learns how ego-motion dynamically modulates pedestrian movement. This allows for accurate inference of future pedestrian states from the egocentric viewpoint. 

Given the superior performance of the Mamba model over GRU, LSTM, and Transformer networks in various sequence prediction tasks, we adopt it as our decoder. In our design, pedestrian motion features serve as the historical context for the decoder, while the vehicle's motion features at the last observed timestamp ($T$) are used as explicit conditional guidance for the future prediction horizon. By leveraging the structured modeling mechanism of the Mamba network, our approach explicitly learns how ego-motion dynamically modulates pedestrian movement. This enables effective inference of future pedestrian states from an egocentric perspective, ultimately outputting the decoded features for future time steps:
\begin{equation}
	F_{de}=Mamba(F_{in})
	\label{eq:7}
\end{equation}
where the input feature sequence $F_{in}$ is composed of features sequence $F_{m}$ and $T_{pred}$ copies of the ego-motion feature at timestamp $T_pred$. Through the decoding operation, a set of decoded features $F_{de}$ that represent the future motion of the pedestrian is captured.

\subsection{Future Trajectory Generator}
The decoder Mamba model captures a set of decoded features $F_{de}^{t:t+n}$ that represent the future motion of a pedestrian. Next, an MLP acts as a generator to map the decoded features of future timestamps into a set of future trajectories. The future trajectory generator is formulated as follows:
\begin{equation}
	\hat{F}_{future}=MLP(F_{de}^{t:t+n})
	\label{eq:8}
\end{equation}

\section{Experiment}
\subsection{Datasets and Metrics}
We conduct experiments to train and evaluate the prediction performance of the proposed model on the widely used benchmark datasets, namely JAAD \cite{Rasouli2017Are} and PIE \cite{Rasouli2019PIE}. JAAD is a dataset for studying joint attention in the context of autonomous driving, which is widely used in trajectory prediction from egocentric views and crossing intention prediction research. To this end, JAAD dataset provides a richly annotated collection of 346 short video clips (5-10 sec long) extracted from over 240 hours of driving footage. PIE is a dataset for studying pedestrian behavior in traffic, which contains over 6 hours of footage recorded in typical traffic scenes with on-board camera. It also provides accurate vehicle information from OBD sensor (vehicle speed, heading direction and GPS coordinates) synchronized with video footage. Following the benchmark protocol, the datasets are randomly split into training (50\%), validation (10\%), and test (40\%) sets. 

Similar to existing works \cite{Rasouli2021Bifold,Rasouli2023PedFormer,Zhang2024Decouple}, we assess the performance of trajectory prediction by using four metrics: Average Displacement Error (ADE), Final Displacement Error (FDE), Average Root Mean Square Error (ARB), and Final Root Mean Square Error (FRB). ADE and FDE measure the displacement error of center point coordinates between the predicted bounding boxes and the ground truth bounding boxes, and ARB and FRB measure the mean square error between the predicted bounding boxes and the coordinates of the groundtruth bounding boxes.

\subsection{Experimental Setup}
We train and evaluate the TrajMamba model on an Nvidia GTX 4080 GPU with CUDA 11.6 and PyTorch 2.0.0. The dimensions of pedestrian motion features and ego-motion features are set to 256. The past and future timestamps are set to 15 (0.5s) and 45 (1.5s). We use the Adam optimizer with an initial learning rate of 0.0001 and smooth $\mathbf{L}_{1}$ loss as the loss function. In particular, for the PIE dataset, the vehicle speed is used to represent the vehicle's movement information to capture ego-motion features, while in the JAAD dataset, the vehicle behaviors (0: stopped, 1: moving slow, 2: moving fast, 3: decelerating, and 4: accelerating) are used as the movement information.

\subsection{Quantitative Evaluation}
We evaluated the proposed TrajMamba model against eight existing methods, including FOL \cite{Yao2019Egocentric}, FPL \cite{Yagi2018Future}, B-LSTM \cite{Bhattacharyya2018Long}, and two variations of the methods in the PIE dataset \cite{Rasouli2019PIE}, BiPed \cite{Rasouli2021Bifold}, PedFormer\cite{Rasouli2023PedFormer}, and two general models BiTrap \cite{Yao2021BiTrap} and SGNet \cite{Wang2022Stepwise}, PEvT \cite{Neumann2021Pedestrian}, MTN \cite{Yin2021Multimodal}, and the model proposed by Zhang et al. \cite{Zhang2024Decouple}. 

The experimental results between TrajMamba and baseline methods on the PIE and JAAD datasets are presented in Table \ref{tab:1}. For 1s prediction (upper part), TrajMamba consistently outperforms all methods. Notably, it surpasses the previous best model PedFormer by 40.98\%/ 44.02\%/ 24.03\%/ 32.91\% on ADE/ FDE/ ARB/ FRB on PIE, and achieves 22.30\%/ 26.11\%/ 26.55\%/ 28.74\% improvements on JAAD. Against general approaches BiTrap and SGNet, TrajMamba yields gains of 12.57\%/ 8.26\%/ 8.01\%/ 4.22\% and 7.54\%/ 4.71\%/ 5.84\%/ 2.57\% on PIE, and 26.92\%/ 21.69\%/ 20.07\%/ 14.90\% and 6.71\%/ 0.39\%/ 5.99\%/ 0.03\% on JAAD, respectively. For 1.5s prediction (lower part), TrajMamba achieves best performance on PIE but lags behind PEvT, MTN, and Zhang et al.'s method on JAAD. This discrepancy likely stems from differences in scene characteristics and vehicle motion representations (speed vs. behavior labels) between datasets.

\subsection{Ablation Study}

\begin{table}[tbp]
	\centering
	\caption{Ablation results on input and output representation.}
	\begin{tabular}{cccccc}
		\toprule
		\multirow{2}[4]{*}{Input} & \multirow{2}[4]{*}{Output} & ADE   & FDE   & ARB   & FRB \\
		\cmidrule{3-6}          &       & \multicolumn{4}{c}{PIE} \\
		\midrule
		LT-BR & LT-BR & 8.72  & 18.69 & 24.68 & 27.88 \\
		xywh  & xywh  & 13.64 & 24.84 & 51.50  & 78.76 \\
		LT-BR + v & LT-BR & 8.58  & 18.20  & 13.15 & 24.16 \\
		xywh+v+s & CV-CS-Offset & \textbf{7.72} & \textbf{16.99} & \textbf{11.60} & \textbf{22.00} \\
		\midrule
		\multicolumn{2}{c}{} & \multicolumn{4}{c}{JAAD} \\
		\midrule
		LT-BR & LT-BR & 17.27 & 35.17 & 23.24 & 41.95 \\
		xywh  & xywh  & 26.66 & 42.12 & 52.21 & 79.19 \\
		LT-BR + v & LT-BR & 15.93 & 32.76 & 20.74 & 38.15 \\
		xywh+v+s & CV-CS-Offset & \textbf{13.90} & \textbf{30.76} & \textbf{18.04} & \textbf{34.79} \\
		\bottomrule
	\end{tabular}%
	\label{tab:2}%
\end{table}%
\textbf{The ablation study on input-output representation:} To evaluate the effectiveness of the proposed input-output representation, we conducted an ablation study, with results reported in Table \ref{tab:2}. In the table, LT-BR denotes the coordinates of the top-left and bottom-right corners of the bounding box; xywh represents the center coordinates along with the width and height; v indicates the velocity along x and y directions; s refers to the scale variation on weight and height; and CV-CS-Offset corresponds to the residual offset derived from the constant velocity and constant scale assumption. Experiment results demonstrate that incorporating velocity information significantly improves prediction accuracy, highlighting the critical role of encoding motion dynamics. Furthermore, the representation that jointly encodes position, velocity, and scale variation—combined with the offset output derived from the CV-CS assumption—further enhances trajectory smoothness and physical plausibility, providing a more robust and interpretable motion feature foundation for the model.

\textbf{The effect of ego-motion-guided decoder:} To verify the effectiveness of the proposed ego-motion-guided decoder, we conducted multiple sets of comparative experiments, with results summarized in Table \ref{tab:3}. For each dataset, we designed two categories of ablation experiments. The first category follows the classic architecture: pedestrian motion features and ego-motion features are extracted separately by encoders, then fused and processed by a decoder to generate future trajectories. The second category corresponds to our proposed architecture, where pedestrian motion features and ego-motion features are extracted separately and then fed into the designed ego-motion-guided decoder to generate future trajectories . Specifically, within each category, we employ LSTM, GRU, Transformer, and Mamba as the encoder/decoder backbones for comprehensive comparison. 

\begin{table}[t]
	\caption{Ablation Results on Ego-motion-guided decoder. (PFD: post-fusion decoding mechanism, EMGD: ego-motion-guided decoding mechanism)}
	\resizebox{1.0\columnwidth}{!}{
	\begin{tabular}{cccccccc}
		\toprule
		& Model & ADE   & FDE   & ARB   & FRB   & Flops & Params \\
		\midrule
		\multicolumn{8}{c}{PIE} \\
		\midrule
		\multirow{4}[1]{*}{PFD} & LSTM  & 7.98  & \textbf{17.82} & 12.14 & 23.29 & 128.27G & 3.16M \\
		& GRU   & 7.97  & 17.98 & 12.25 & 23.72 & 100.68G & 2.51M \\
		& Transformer & 7.94  & 17.86 & 12.25 & 23.52 & 87.48G & 2.31M \\
		& Mamba & \textbf{7.91} & 17.87 & \textbf{11.99} & \textbf{23.19} & \textbf{30.11G} & \textbf{1.32M} \\
		\midrule
		\multirow{4}[1]{*}{EMGD} & LSTM  & 8.14  & 18.26 & 12.34 & 23.57 & 63.64G & 1.78M \\
		& GRU   & 7.97  & 17.87 & 12.13 & 23.28 & 49.34G & 1.39M \\
		& Transformer & \textbf{7.72} & 17.13 & 11.84 & 22.51 & 6.35G & 0.20M \\
		& Mamba & \textbf{7.72} & \textbf{16.99} & \textbf{11.6} & \textbf{22} & \textbf{6.35G} & \textbf{0.20M} \\
		\midrule
		\multicolumn{8}{c}{JAAD} \\
		\midrule
		\multirow{4}[1]{*}{PFD} & LSTM  & 15.66 & 35.37 & 20.16 & 39.07 & 220.99G & 3.16M \\
		& GRU   & 15.19 & 34.51 & 19.71 & 38.29 & 173.46G & 2.51M \\
		& Transformer & 17.12 & 37.87 & 21.04 & 39.34 & 150.71G & 2.31M \\
		& Mamba & \textbf{14.34} & \textbf{32.43} & \textbf{18.37} & \textbf{35.63} & \textbf{51.86G} & \textbf{1.32M} \\
		\midrule
		\multirow{4}[1]{*}{EMGD} & LSTM  & 14.86 & 32.91 & 19.28 & 36.76 & 196.82G & 1.78M \\
		& GRU   & 14.46 & 31.91 & 18.76 & 35.74 & 85.00G & 1.39M \\
		& Transformer & 14.05 & 31.38 & 18.23 & 35.44 & 10.94G & 0.20M \\
		& Mamba & \textbf{13.9} & \textbf{30.76} & \textbf{18.04} & \textbf{34.79} & \textbf{10.94G} & \textbf{0.20M} \\
		\bottomrule
	\end{tabular}%
}
	\label{tab:3}%
\end{table}%
From the results presented in the table, it can be observed that for the same base model (with the exception of LSTM), the variant constructed with our proposed ego-motion-guided decoding mechanism (EMGD) consistently achieves more accurate predictions than the classic post-fusion decoding mechanism (PFD). This improvement arises because traditional methods provide a fused motion representation that mixes pedestrian motion with ego-motion during decoding, leading to ambiguity and making it difficult for the model to disentangle the relative motion relationship between the pedestrian and the vehicle. In contrast, our proposed ego-motion-guided decoder takes pedestrian motion features as historical context and ego-motion features as conditioning information for future steps, explicitly guiding the decoder to capture the dynamic regulatory mechanism of ego-motion on pedestrian motion. This enables more accurate inference of pedestrian motion states from an ego-centric perspective. On the other hand, across all comparative experiments, models utilizing Mamba as the backbone network consistently achieve the smallest parameter counts and lowest computational costs, regardless of the decoding mechanism employed. Furthermore, models built with the proposed ego-motion-guided decoding (EMGD) mechanism generally exhibit fewer parameters and reduced computational requirements. 

\textbf{The effect of Mamba encoder:} To further investigate the critical role of the Mamba network in motion feature extraction, we design targeted ablation experiments. In previous works, LSTM, GRU, and Transformer have been the three most prevalent sequence models for extracting pedestrian motion features and vehicle ego-motion features. In contrast, our proposed TrajMamba model adopts Mamba as the encoder to extract both types of motion features. Accordingly, we design four experimental configurations: with the Mamba network fixed as the decoder, we employ LSTM, GRU, Transformer, and Mamba as the encoders, respectively. Ablation studies are conducted on both the PIE and JAAD datasets, with the results presented in Table \ref{tab:4}. The experimental results demonstrate that using Mamba as the encoder achieves significantly better prediction performance (i.e., lower errors) across the four metrics of ADE, FDE, ARB, and FRB compared to the other three models. This indicates that the structured state-space design of the Mamba model is more effective than recurrent neural networks and attention-based Transformer networks in extracting motion features through temporal encoding.

\begin{table}[t]
	\centering
	\caption{Ablation Results on Encoder.}
	\begin{tabular}{cccccc}
		\toprule
		\multicolumn{1}{c}{\multirow{2}[4]{*}{Encoder}} & \multicolumn{1}{c}{\multirow{2}[4]{*}{Decoder}} & \multicolumn{1}{l}{ADE} & \multicolumn{1}{l}{FDE} & \multicolumn{1}{l}{ARB} & \multicolumn{1}{l}{FRB} \\
		\cmidrule{3-6}          &       & \multicolumn{4}{c}{PIE} \\
		\midrule
		LSTM  & Mamba & 8.13  & 18.23 & 12.28 & 23.58 \\
		GRU   & Mamba & 7.87  & 17.41  & 11.89 & 22.67 \\
		Transformer & Mamba & 7.81  & 17.46 & 11.84  & 22.67 \\
		\midrule
		\midrule
		Mamba & Mamba & \textbf{7.72}  & \textbf{16.99} & \textbf{11.60} & \textbf{22.00} \\
		\midrule
		&       & \multicolumn{4}{c}{JAAD} \\
		\midrule
		LSTM  & Mamba & 14.49 & 32.29 & 18.93 & 36.32 \\
		GRU   & Mamba & 14.00 & 31.13 & 18.16 & 35.00 \\
		Transformer & Mamba & 14.38 & 32.37 & 18.88 & 36.89 \\
		\midrule
		\midrule
		Mamba & Mamba & \textbf{13.90} & \textbf{30.76} & \textbf{18.04} & \textbf{34.79} \\
		\bottomrule
	\end{tabular}%
	\label{tab:4}%
\end{table}%

\textbf{The effect of Mamba decoder:} To further validate the critical role of Mamba as a ego-motion-guided decoder, we designed additional ablation experiments. Specifically, we fixed Mamba as the encoder and varied the decoder architecture, employing LSTM, GRU, Transformer, and Mamba as decoders, respectively. The experimental results are presented in Table \ref{tab:5}. The results demonstrate that Mamba consistently achieves the best prediction performance as the decoder on both datasets, followed by Transformer as the second-best. These results demonstrate that Mamba is effective not only as a motion encoder but also as a decoder for modeling the regulatory mechanism of ego-motion on pedestrian movement.

\begin{table}[t]
	\centering
	\caption{Ablation Results on Decoder.}
	\begin{tabular}{cccccc}
		\toprule
		\multicolumn{1}{c}{\multirow{2}[4]{*}{Encoder}} & \multicolumn{1}{c}{\multirow{2}[4]{*}{Decoder}} & \multicolumn{1}{l}{ADE} & \multicolumn{1}{l}{FDE} & \multicolumn{1}{l}{ARB} & \multicolumn{1}{l}{FRB} \\
		\cmidrule{3-6}          &       & \multicolumn{4}{c}{PIE} \\
		\midrule
		Mamba  & LSTM & 8.05  & 18.01 & 12.56 & 24.03 \\
		Mamba  & GRU & 7.98  & 17.90  & 12.33 & 23.59 \\
		Mamba & Transformer & 7.72  & 17.33 & 11.90  & 22.87 \\
		\midrule
		\midrule
		Mamba & Mamba & \textbf{7.72}  & \textbf{16.99} & \textbf{11.60} & \textbf{22.00} \\
		\midrule
		&       & \multicolumn{4}{c}{JAAD} \\
		\midrule
		Mamba  & LSTM & 14.54 & 32.44 & 18.95 & 36.40 \\
		Mamba  & GRU  & 14.75 & 32.65 & 18.97 & 36.15 \\
		Mamba & Transformer & 14.06 & 31.58 & 18.31 & 35.42 \\
		\midrule
		\midrule
		Mamba & Mamba & \textbf{13.90} & \textbf{30.76} & \textbf{18.04} & \textbf{34.79} \\
		\bottomrule
	\end{tabular}%
	\label{tab:5}%
\end{table}%

\begin{figure*}[htbp]
	\centering
	\begin{minipage}{0.22\linewidth}0s
		\centering
		\includegraphics[width=1\linewidth]{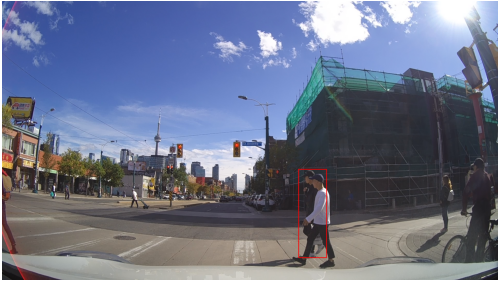}
		\quad
	\end{minipage}
	\begin{minipage}{0.22\linewidth}+0.5s
		\centering
		\includegraphics[width=1\linewidth]{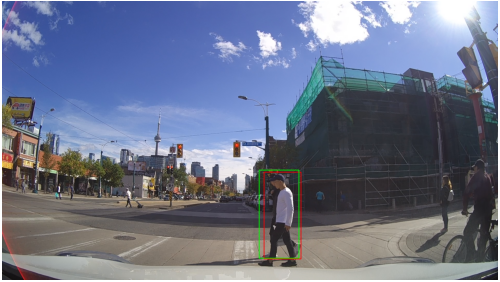}
		\quad
	\end{minipage}
	\begin{minipage}{0.22\linewidth}+1.0s
		\centering
		\includegraphics[width=1\linewidth]{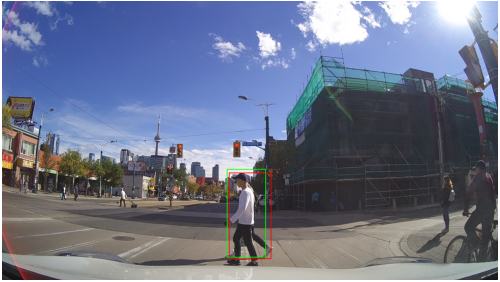}
		\quad
	\end{minipage}
	\begin{minipage}{0.22\linewidth}+1.5s
		\centering
		\includegraphics[width=1\linewidth]{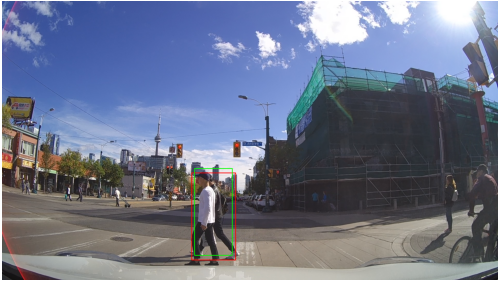}
		\quad
	\end{minipage}
	\begin{minipage}{0.22\linewidth}
		\centering
		\includegraphics[width=1\linewidth]{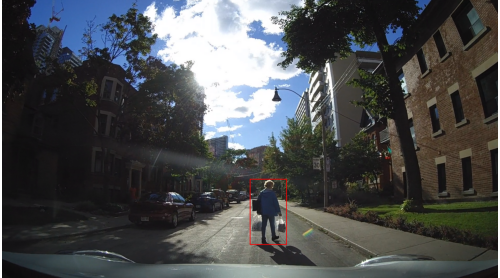}
		\quad
	\end{minipage}
	\begin{minipage}{0.22\linewidth}
		\centering
		\includegraphics[width=1\linewidth]{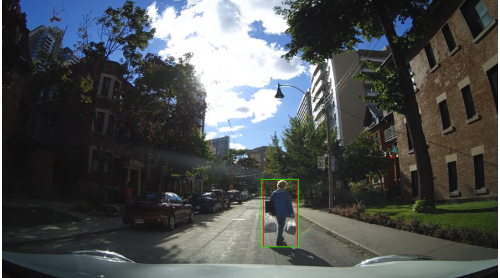}
		\quad
	\end{minipage}
	\begin{minipage}{0.22\linewidth}
		\centering
		\includegraphics[width=1\linewidth]{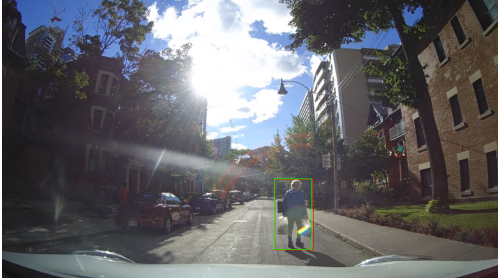}
		\quad
	\end{minipage}
	\begin{minipage}{0.22\linewidth}
		\centering
		\includegraphics[width=1\linewidth]{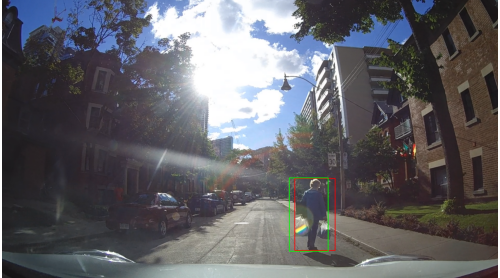}
		\quad
	\end{minipage}
	
	\begin{minipage}{0.22\linewidth}
		\centering
		\includegraphics[width=1\linewidth]{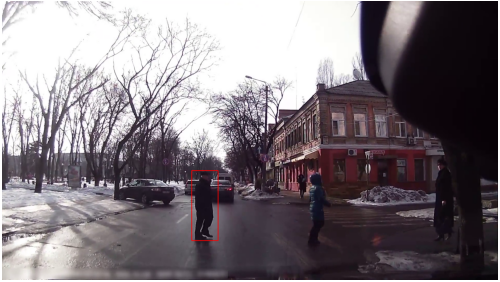}
		\quad
	\end{minipage}
	\begin{minipage}{0.22\linewidth}
		\centering
		\includegraphics[width=1\linewidth]{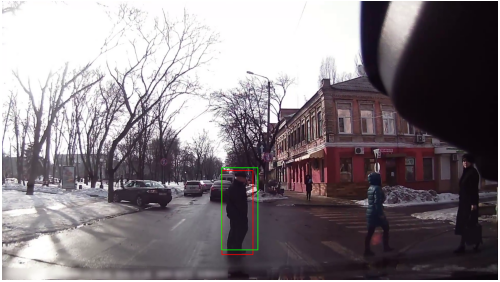}
		\quad
	\end{minipage}
	\begin{minipage}{0.22\linewidth}
		\centering
		\includegraphics[width=1\linewidth]{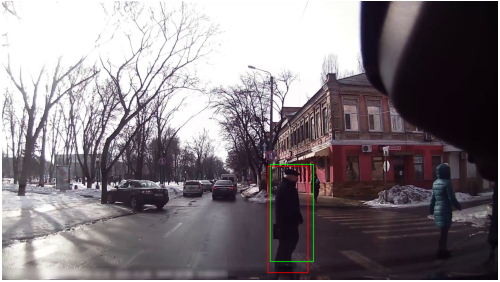}
		\quad
	\end{minipage}
	\begin{minipage}{0.22\linewidth}
		\centering
		\includegraphics[width=1\linewidth]{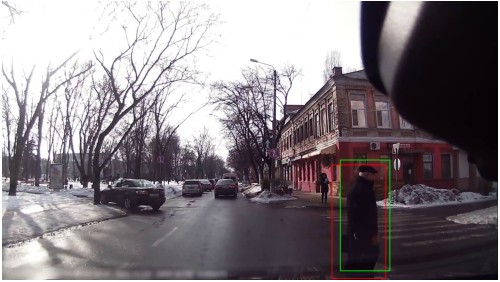}
		\quad
	\end{minipage}
	
	\begin{minipage}{0.22\linewidth}
		\centering
		\includegraphics[width=1\linewidth]{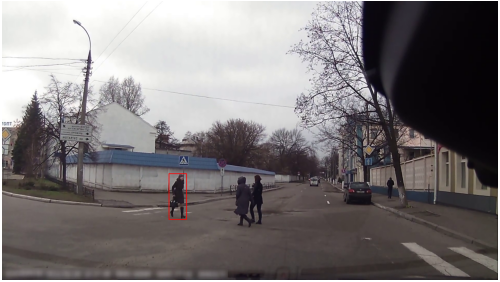}
	\end{minipage}
	\begin{minipage}{0.22\linewidth}
		\centering
		\includegraphics[width=1\linewidth]{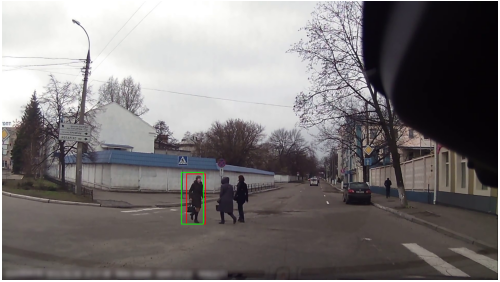}
	\end{minipage}
	\begin{minipage}{0.22\linewidth}
		\centering
		\includegraphics[width=1\linewidth]{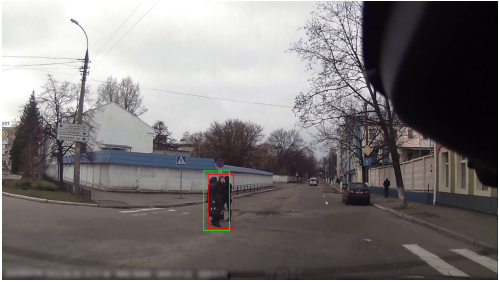}
	\end{minipage}
	\begin{minipage}{0.22\linewidth}
		\centering
		\includegraphics[width=1\linewidth]{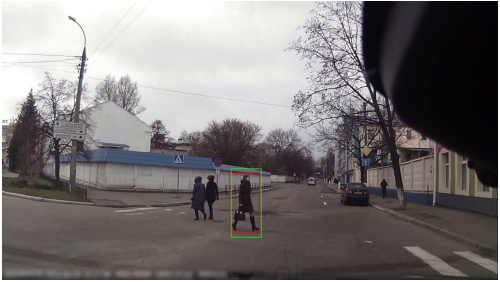}
	\end{minipage}
	\caption{Pedestrian trajectory prediction qualitative samples on the PIE and JAAD datasets. Ground truth positions in red, predictions in green. Best viewed in colour.}
	\label{fig:3}
\end{figure*}

\subsection{Qualitative Evaluation}
In this section, we visually demonstrate the effectiveness of the proposed TrajMamba model through qualitative evaluation results. Some visualization results of prediction samples of TrajMamba on the PIE and JAAD datasets are shown in Figure \ref{fig:3}. The first column shows the position of the pedestrian in the last observed frame, and the second to fourth columns, respectively, show the future positions at 0.5s, 1.0s, and 1.5s. The top two rows are samples from the PIE dataset, and the bottom two rows are from the JAAD dataset. These visualization results indicate that TrajMamba can effectively model the motion state of pedestrians and ego-motion and accurately predict the future trajectory of the tracked pedestrian from the egocentric perspective.

\section{Conclusion}
We propose a novel Mamba-based model called TrajMamba to predict the future trajectory of the tracked pedestrian from an egocentric perspective. TrajMamba innovatively employs the Mamba model to extract pedestrian motion features and the ego-motion of the ego-camera. Furthermore, an ego-motion-guided Mamba model is proposed to jointly model the relative relationship between pedestrian movement and camera movement through sequential modeling and effectively infer the future movement of pedestrians, thereby accurately predicting the future trajectory of pedestrians from the perspective of ego-motion. Experimental results on PIE and JAAD datasets validate the effectiveness and excellent performance of TrajMamba. Currently, we have not yet considered the mutual influence among pedestrians in crowds. Therefore, in future work, the prediction performance of the TrajMamba model will be improved by modeling pedestrian interaction.

\bibliographystyle{IEEEtran}
\bibliography{IEEEfull}

\end{document}